\documentclass[conference,a4paper]{IEEEtran}
\IEEEoverridecommandlockouts

\usepackage{cite}
\usepackage{amsmath,amssymb,amsfonts}
\usepackage{algorithmic}
\usepackage{graphicx}
\usepackage{textcomp}
\usepackage{xcolor}
\usepackage{booktabs}
\usepackage{makecell}
\usepackage{multirow}

\def\BibTeX{{\rm B\kern-.05em{\sc i\kern-.025em b}\kern-.08em
    T\kern-.1667em\lower.7ex\hbox{E}\kern-.125emX}}
\begin{document}

\title{Physics-Informed Operator Learning for Hemodynamic Modeling}

\author{
\IEEEauthorblockN{Ryan Chappell}
\IEEEauthorblockA{\textit{School of Computing and Electrical Engineering} \\
\textit{Queensland University of Technology}\\
Brisbane, Australia \\
ryan.chappell@connect.qut.edu.au}
\and
\IEEEauthorblockN{Chayan Banerjee\textsuperscript{*}}
\IEEEauthorblockA{\textit{School of Computing and Electrical Engineering} \\
\textit{Queensland University of Technology}\\
Brisbane, Australia \\
c.banerjee@qut.edu.au}
\and
\IEEEauthorblockN{Kien Nguyen}
\IEEEauthorblockA{\textit{School of Computing and Electrical Engineering} \\
\textit{Queensland University of Technology}\\
Brisbane, Australia \\
k.nguyenthanh@qut.edu.au}
\and
\IEEEauthorblockN{Clinton Fookes}
\IEEEauthorblockA{\textit{School of Computing and Electrical Engineering} \\
\textit{Queensland University of Technology}\\
Brisbane, Australia \\
c.fookes@qut.edu.au}
\thanks{\textsuperscript{*}Corresponding author.}
}

\maketitle

\begin{abstract}
Accurate modeling of personalized cardiovascular dynamics is crucial for non-invasive monitoring and therapy planning. State-of-the-art physics-informed neural network (PINN) approaches employ deep, multi-branch architectures with adversarial or contrastive objectives to enforce partial differential equation constraints. While effective, these enhancements introduce significant training and implementation complexity, limiting scalability and practical deployment.
We investigate physics-informed neural operator learning models as efficient supervisory signals for training simplified architectures through knowledge distillation. Our approach pre-trains a physics-informed DeepONet (PI-DeepONet) on high-fidelity cuffless blood pressure recordings to learn operator mappings from raw wearable waveforms to beat-to-beat pressure signals under embedded physics constraints. This pre-trained operator serves as a frozen supervisor in a lightweight knowledge-distillation pipeline, guiding streamlined base models that eliminate complex adversarial and contrastive learning components while maintaining performance.
We characterize the role of physics-informed regularization in operator learning and demonstrate its effectiveness for supervisory guidance. Through extensive experiments, our operator-supervised approach achieves performance parity with complex baselines (correlation: 0.766 vs. 0.770, RMSE: 4.452 vs. 4.501), while dramatically reducing architectural complexity from eight critical hyperparameters to a single regularization coefficient and decreasing training overhead by 4\% . Our results demonstrate that operator-based supervision effectively replaces intricate multi-component training strategies, offering a more scalable and interpretable approach to physiological modeling with reduced implementation burden. 
\end{abstract}

\begin{IEEEkeywords}
Physics-Informed Neural Networks, 
Neural Operator Learning,
DeepONet,
Cuffless Blood Pressure Estimation
\end{IEEEkeywords}

\section{Introduction}
Modeling physiological systems such as hemodynamics often requires balancing physical accuracy, computational efficiency, and generalizability. Physics-informed neural networks (PINNs) have become a prominent approach for incorporating domain knowledge into neural network models. However, PINNs face well-documented challenges: inefficiencies and inaccuracies when solving differential equations over long temporal horizons, poor enforcement of conservation laws, and high training costs compared to traditional numerical solvers \cite{Wang2023,Brecht2024}.

Deep operator networks (DeepONets), in contrast, approximate solution operators rather than solutions themselves. By learning mappings between function spaces, DeepONets achieve improved generalization and offer architectural simplicity \cite{Lu2021}. Their structure—comprising a branch network for encoding input functions and a trunk network for representing output locations—makes them inherently suited to model spatiotemporal dynamics, a critical need in physiological modeling. However, standard DeepONets require extensive supervised datasets, which can be limiting in personalized healthcare applications.

To address this, physics-informed DeepONets (PI-DeepONets) incorporate governing physical principles as constraints during training, reducing data requirements while maintaining accuracy \cite{Wang2021}. These features position PI-DeepONets as a promising solution for personalized, long-horizon modeling tasks like continuous blood pressure monitoring. In this work, we explore whether a physics-informed operator learning approach can offer a simpler, more efficient alternative to existing PINN-based architectures—preserving accuracy while reducing complexity, and ultimately enabling more scalable and interpretable physiological modeling.

\begin{figure*}[t]
    \centering
    \includegraphics[width=0.9\textwidth]{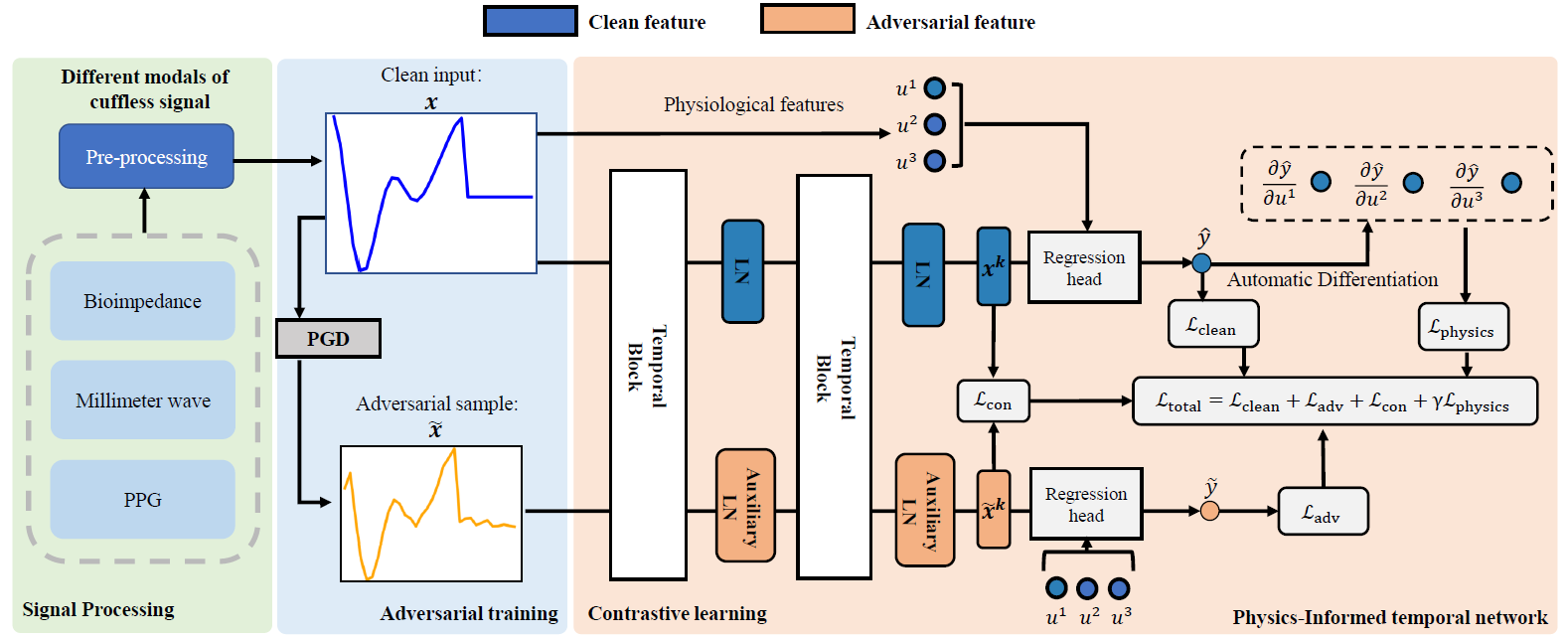}
    \caption{PITN-Full approach\cite{wang2024adversarial}, the overall framework comprises two main data flows: one for clean input signals used for prediction and Taylor-based differentiation, and another for adversarial samples generated via PGD. Adversarial training and a contrastive module are incorporated to enhance learning under limited physiological data. During inference, auxiliary layer normalizations (LNs) are disabled to account for distributional differences between clean and adversarial inputs. }
    \label{fig:PITN-Full}
\end{figure*}

\section{Related Work}
\noindent
\textit{Physics-Informed Neural Networks in Healthcare :} 
PINNs have been successfully applied across diverse domains, including computer vision \cite{banerjee2024physics}, reinforcement learning \cite{banerjee2025survey}, medical image analysis \cite{banerjee2024pinns}, and hemodynamic modeling \cite{wang2024adversarial}. While effective for incorporating domain knowledge, PINNs face well-documented challenges in physiological applications: inefficiencies over long temporal horizons, poor conservation law enforcement, and high training costs compared to traditional numerical solvers.

In hemodynamic modeling specifically, \cite{kissas2020machine} demonstrated that PINNs can predict arterial blood flow dynamics from sparse clinical data, outperforming traditional computational fluid dynamics (CFD) simulations in accuracy and efficiency. \cite{herrero2022ep} applied PINNs to simulate cardiac electrophysiology, illustrating their capacity to capture intricate physiological phenomena governed by nonlinear PDEs. However, these applications reveal fundamental limitations: numerical inaccuracies accumulate over time, leading to significant deviations from true solutions \cite{Cardoso-Bihlo2023}. The high-dimensional, non-convex optimization landscape often results in convergence to suboptimal local minima, especially for stiff cardiovascular systems \cite{DeRyck2022}. The reliance on pointwise residuals also requires highly regular solutions, limiting applicability to hemodynamic PDEs with discontinuities or sharp pressure gradients \cite{Ernst2024, Sharma2023}.

\noindent
\textit{Conventional Methods and Limitations :}
Traditional computational methods in hemodynamic modelling, such as CFD simulations, have supported high-fidelity modelling of arterial blood flow and valve dynamics using finite element methods for pressure wave propagation. However, these methods have suffered from significant computational cost and challenges in handling complex geometric structures \cite{Taebi2022}, \cite{Xie2025}. Further, clinical cuff-based blood pressure monitoring remains discrete, intermittent and uncomfortable 
\cite{Zhao2023} . Such limitations have motivated the adoption of continuous, cuffless measurement techniques for blood pressure estimation using wearable technologies, including bioimpedance, photoplethysmography (PPG), and mmWave radar \cite{Hua2024}. 

Early data-driven approaches using traditional machine learning methods and signal processing have demonstrated poor generalisation and susceptibility to domain shifts, highlighting the opportunity for integration of physiological constraints \cite{Shen2025}. Moreover, variability of cardiovascular responses among individuals presents challenges in clinical settings which require patient-specific modeling \cite{Sel2024}.
These limitations have spurred the development of physics-informed neural networks (PINNs), which embed physiological laws into the learning process, though their complexity remains a barrier to clinical adoption

To address challenges in physiological time-series analysis, recent methods like PITN \cite{wang2024adversarial} use temporal blocks, adversarial training, and contrastive learning for cuffless blood pressure estimation. Although effective, these add substantial computational complexity and training overhead, limiting clinical scalability. 

\noindent
\textit{Neural Operator Learning as an Alternative :} 
Neural Operator Learning \cite{kovachki2023neural} offers a compelling alternative framework for modeling complex cardiovascular systems. Unlike PINNs, which enforce differential constraints pointwise and struggle with long-term hemodynamic accuracy, operator learning approximates entire mappings from input physiological signals to output pressure waveforms in a function space-aware manner \cite{li2024physics}. DeepONet \cite{lu2019deeponet} learns nonlinear operators by decomposing the problem into branch and trunk networks, generalizing across function inputs and capturing global operator behavior more efficiently than pointwise-residual methods.

\begin{figure*}[t]
    \centering
    \includegraphics[width=\textwidth]{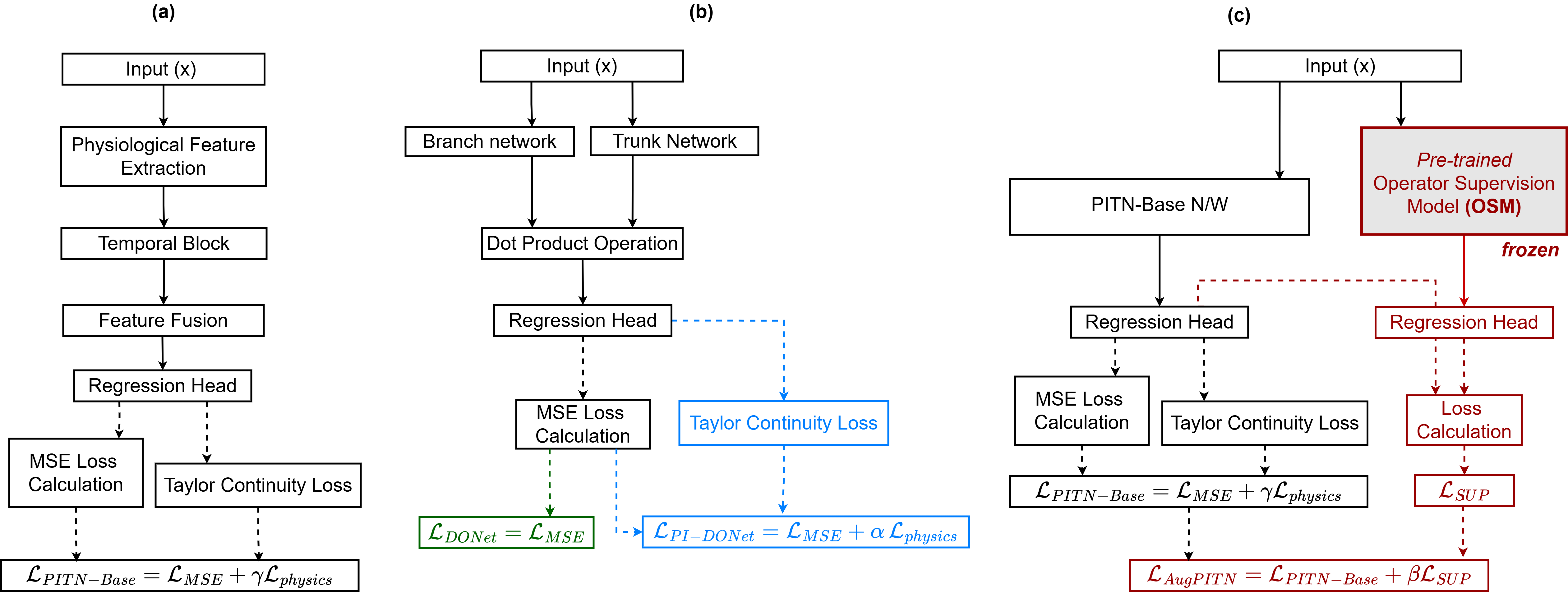}
    \caption{Experimented approaches (a) PITN-Base: simplified version of the PITN-Full, without the Adversarial and Contrastive learning augmentations (b) DeepONet and PI-DeepONet (additional part is shown in black): Neural Operator Learning models (c) The proposed Augmented PITN architecture \textbf{(AugPITN)}: PITN-Base + Operator Supervision Model (OSM): our proposed augmentation model using a pretrained Neural Operator Model as a supervisor to train the base version of PITN (i.e. PITN-Base)}
    \label{fig:arch}
\end{figure*}

This offers greater robustness to sharp pressure gradients and discontinuities by learning the solution operator directly, without enforcing PDE constraints during training.
PI-DeepONet \cite{Wang2021} extends this framework by incorporating physics-based constraints through PDE violation penalties during training. By combining operator learning generalization with physics-informed losses, it enhances accuracy and consistency in data-scarce physiological regimes while retaining DeepONet's real-time inference efficiency—critical for continuous cardiovascular monitoring applications.

While PINNs show promise in hemodynamic modeling, they struggle with high computational costs, convergence issues, and weak conservation over long time horizons. More complex variants like PITN use adversarial and contrastive learning to improve results but introduce heavy implementation and tuning burdens that hinder clinical adoption. In contrast, DeepONets offer architectural simplicity and better generalization by learning mappings between function spaces. Once trained, they allow fast, real-time inference without retraining. This work explores whether physics-informed operator learning can serve as an efficient supervisory signal via knowledge distillation, matching the performance of complex adversarial models with lower computational and training costs suitable for personalized cardiovascular applications.

\section{Problem Formulation and Contributions}
The Physics-Informed Temporal Network (PITN) was developed for personalized hemodynamic modeling using bioimpedance signals \cite{wang2024adversarial}. It combines a temporal convolutional block that reshapes time series into 2D for feature extraction, a personalized regression head conditioned on physiological parameters, and a physics-based loss derived from a Taylor series approximation. It further integrates adversarial and contrastive learning to improve generalization and representation quality. While effective, these components introduce substantial complexity and computational burden (see Fig.~\ref{fig:PITN-Full}). \newline
\textcolor{black}{We investigate whether a PI-DeepONet-based model can replace PITN as a simpler, more efficient alternative, preserving accuracy while reducing training and architectural overhead. The key scalability gain comes from hyperparameter reduction: PITN-Full requires tuning eight coupled parameters (adversarial strength $\eta$, contrastive temperature $\tau$, positive pair threshold $y_{\text{shift}}$, physics weight $\gamma$, perturbation bound $\epsilon$, and distinct learning rates for clean/adversarial paths), whereas our operator-supervision framework collapses this to a single hyperparameter $\beta$ controlling supervision strength. This eliminates costly multi-dimensional grid searches and simplifies implementation from managing dual clean/adversarial data flows to a standard supervised pipeline with frozen operator guidance.}

The primary contributions of this work are as follows:
\begin{itemize}
    \item \textit{Simplified Operator-Supervised Architecture: }We propose AugPITN, a lightweight alternative to PITN-Full—a representative of high-complexity physics-informed models that rely on adversarial and contrastive learning. AugPITN replaces these components with supervision from a pre-trained PI-DeepONet model, enabling a significantly simpler architecture without sacrificing performance (see Fig.~\ref{fig:arch}).
    
    \item \textit{Functional Supervision as a Scalable Training Paradigm:}
    We show that operator-based functional supervision can serve as an effective alternative to the intricate training procedures used in PITN-Full. By leveraging a frozen neural operator as a global supervisory signal, our method avoids instance-level objectives and manual tuning, promoting stable and scalable training for learning high-level functional mappings.

\end{itemize}

\section{Methodology}
We propose a novel approach that leverages pre-trained operator models as supervisory signals to improve the training of simpler baseline architectures. This augmentation framework, termed \textbf{AugPITN}, incorporates a generic supervisory mechanism referred to as the \textit{Operator Supervision Model (OSM)}, which incorporates advanced neural operator models—specifically DeepONet and its physics-informed variant PI-DeepONet—to guide the training process of a reduced-complexity base model.

By eliminating the need for deep, multi-branch architectures and complex adversarial or contrastive training, OSM provides a streamlined yet effective supervisory mechanism. Pre-trained on high-fidelity cuffless blood pressure data, the operator models learn physiologically consistent mappings from wearable signals to beat-to-beat pressure under embedded physics constraints. Once trained and frozen, these operators guide the training of a simplified PITN variant through loss-level supervision.

This operator-based augmentation enables efficient knowledge transfer, reducing model complexity and training overhead while maintaining robust prediction capabilities. The framework balances physics-informed regularization with practical model simplicity, forming the core of the AugPITN training approach.

\subsection{Primary Baseline Models }
PITN \cite{wang2024adversarial} is a hybrid deep learning framework developed for cuffless blood pressure (BP) estimation using time-series data from wearable sensors such as bioimpedance (BioZ), photoplethysmography (PPG), and millimeter-wave (mmWave) radar. 
The \textbf{PITN-Base} model integrates temporal modeling with Taylor-based physics-informed regularization, while \textbf{PITN-Full} extends this by incorporating adversarial training and contrastive learning to enhance robustness and generalization.

\textbf{PITN-Base} consists of two core components: a temporal modeling block and a physics-informed loss.\newline
\noindent
\textit{Temporal Block:} Transforms 1D signals into 2D tensors by estimating dominant frequency $f$ via FFT, computing period $p = \lceil T / f \rceil$, and reshaping to $(p, f)$ with padding. Multi-scale 2D Inception convolution extracts temporal features with residual connections: $x_k = \text{TemporalBlock}(x_{k-1}) + x_{k-1}$.

\textit{Physics-Informed Loss:} Uses first-order Taylor approximation:
\begin{align}
\tilde{f}_i(x, u, \theta) &= f(x_i, u_i, \theta) \nonumber \\
&\quad + \nabla_{u_i} f(x_i, u_i, \theta)^T (u - u_i), \\
h_i(x_{i+1}, u_{i+1}, \theta) &= \tilde{f}_i(x_{i+1}, u_{i+1}, \theta) - f(x_{i+1}, u_{i+1}, \theta), \\
\mathcal{L}_{\text{physics}} &= \frac{1}{N - 1} \sum_{i=1}^{N-1} h_i^2.
\end{align}

The complete loss combines regression and physics constraints:
\begin{align}
\mathcal{L}_{\text{PITN-Base}} = \mathcal{L}_{\text{MSE}} + \gamma \mathcal{L}_{\text{physics}}
\end{align}
where $\mathcal{L}_{\text{MSE}} = \frac{1}{S} \sum_{i=1}^{S} (\hat{y}_i - y_i)^2$ and $\gamma = 1$.

\vspace{0.2cm}

\textbf{PITN-Full} enhances PITN-Base with adversarial augmentation and contrastive regularization.

\textit{Adversarial Training:} Projected Gradient Descent (PGD) perturbs inputs as $\tilde{x} = \text{clip}_{\Omega}(x + \Delta)$, with $\Delta = \eta \cdot \text{sign}(\nabla_x f(x, u, \theta))$, encouraging robustness to signal variation.

\textit{Contrastive Learning:} To align similar BP representations, contrastive loss is applied. For each sample $i$, the set of positive samples is defined as $P(i) = \{p : |y_i - y_p| < y_{\text{shift}}\}$, while $A(i)$ represents all samples. The contrastive loss is:
\[
\mathcal{L}_{\text{con}} = \sum_{i \in S} -\frac{1}{|P(i)|} \sum_{p \in P(i)} \log \frac{\exp(x_i \cdot x_p / \tau)}{\sum_{a \in A(i)} \exp(x_i \cdot x_a / \tau)}.
\]
The total loss combines regression and regularization terms:
\[
\mathcal{L}_{\text{total}} = \mathcal{L}_{\text{clean}} + \mathcal{L}_{\text{adv}} + \mathcal{L}_{\text{con}} + \gamma \mathcal{L}_{\text{physics}},
\]
where each term corresponds to clean MSE, adversarial MSE, contrastive alignment, and physics-based consistency, respectively.

\subsection{Neural Operator Models: DeepONet, PI-DeepONet \& Variants }

We use \textbf{DeepONet}~\cite{lu2019deeponet,Lu2021} as a framework for learning nonlinear operators from physiological time-series data. It comprises two subnetworks: the branch network encodes discrete input function values \(u\) at sensor locations \(\{x_1, x_2, \ldots, x_m\}\) into features \(b_1, b_2, \ldots, b_p\), and the trunk network encodes query locations \(y\) into features \(t_1, t_2, \ldots, t_p\). The operator output is computed as

\begin{equation}
G(u)(y) \approx \sum_{k=1}^p b_k(u(x_1), \ldots, u(x_m)) \cdot t_k(y) + b_0,
\end{equation}

where \(b_0\) is a learnable bias. This architecture satisfies the universal approximation theorem for operators:

\begin{equation}
\| G(u)(y) - \text{DeepONet}(u,y) \|_\infty < \varepsilon
\end{equation}

for any continuous operator \(G\) and \(\varepsilon > 0\). DeepONet is trained via mean squared error (MSE):

\begin{equation}
\mathcal{L}_{\text{DONet}} = \frac{1}{N} \sum_{i=1}^N \left[ G(u_i)(y_i) - \text{DeepONet}(u_i, y_i) \right]^2,
\end{equation}

enabling generalization over varying inputs and evaluation points.

Extending this, \textbf{PI-DeepONet} introduces a physics-informed constraint by augmenting the MSE loss:

\begin{equation}
\mathcal{L}_{\text{PI-DONet}} = \mathcal{L}_{\text{MSE}} + \lambda_{\text{physics}} \cdot \mathcal{L}_{\text{physics}}.
\end{equation}

The physics loss enforces temporal consistency via a first-order Taylor approximation:

\begin{align}
\tilde{f}_i &= \text{DeepONet}(u_i, y_i)  \nonumber \\
&\quad + \nabla_u \text{DeepONet}(u_i, y_i)^T (u_{i+1} - u_i) \\
h_i &= \tilde{f}_i - \text{DeepONet}(u_{i+1}, y_{i+1}) \\
\mathcal{L}_{\text{physics}} &= \frac{1}{N-1} \sum_{i=1}^{N-1} h_i^2
\end{align}

where \(\lambda_{\text{physics}}\) balances accuracy and physical consistency without compromising universal approximation.\newline
Within the \textbf{Operator Supervision Model (OSM)} framework, we augment the PITN-Base loss with supervision from a pre-trained (frozen) operator model, DeepONet or PI-DeepONet:

\begin{equation}
\mathcal{L}_{\text{AugPITN}} = \mathcal{L}_{\text{PITN-Base}} + \beta \cdot \mathcal{L}_{\text{SUP}},
\end{equation}

where \(\beta\) controls supervision strength.

Two supervision strategies are used:
\begin{itemize}
    \item \textbf{AugPITN (A): Ground Truth-Referenced Supervision} — The OSM output is supervised against the ground truth using \(\mathcal{L}_{\text{SUP}} = \mathrm{MSE}\left( \mathrm{OSM}(u),\ y_{\text{true}} \right)\), where \(\mathrm{OSM}(u) \in \{\text{DeepONet}(u), \text{PI\text{-}DeepONet}(u)\}\).
    
    \item \textbf{AugPITN (B): Prediction Alignment with the Operator} — The PITN-Base output aligns with the OSM’s prediction via \(\mathcal{L}_{\text{SUP}} = \mathrm{MSE}\left( f_{\text{PITN-Base}}(u),\ \mathrm{OSM}(u) \right)\).
\end{itemize}

\textcolor{black}{The supervision mechanism in AugPITN (A) operates as follows: while the pre-trained operator OSM is frozen and non-trainable, its outputs $\text{OSM}(u)$ are compared against ground truth $y_{\text{true}}$ to compute $\mathcal{L}_{\text{SUP}}$. During backpropagation, this supervisory loss provides gradients that flow back to update PITN-Base parameters, encouraging the base network to learn representations consistent with the operator's physics-informed functional mapping. The frozen operator acts as a regularizing supervisor, guiding the base network toward physiologically plausible predictions while transferring its inductive biases, thereby improving generalization and training stability, particularly in data-scarce settings.}

\begin{figure*}[t]
    \centering
    \includegraphics[width=0.98\linewidth]{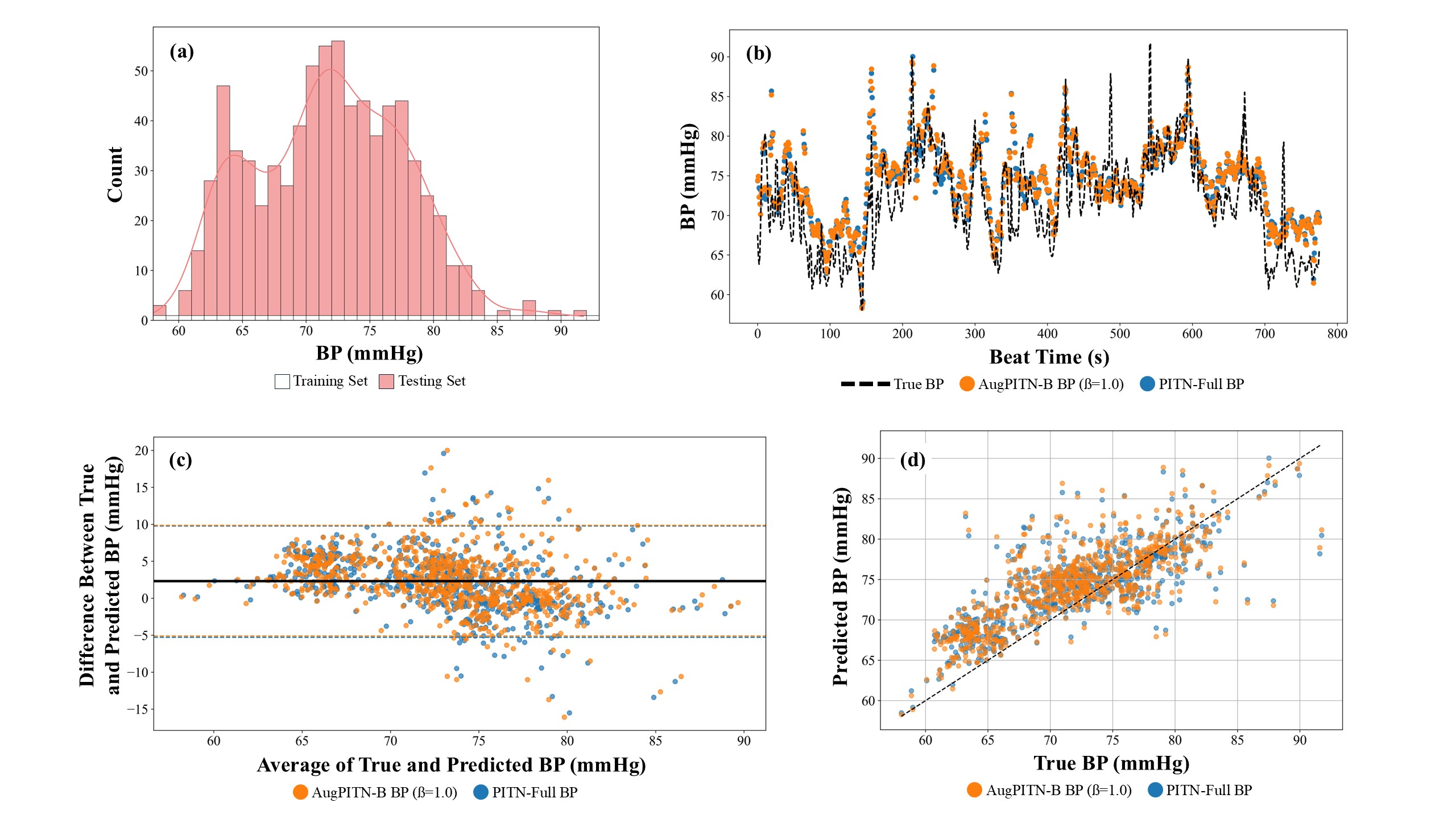}
    \caption{Introducing train/test dataset, baseline (\textbf{PITN-Full}) and best proposed model performance (\textbf{AugPITN (B)}): 
    (a) Histogram showing the distribution of training and testing instances across the pressure range for one subject.
    (b) Beat-to-beat pressure predictions from PITN-Full (black) and AugPITN (B) (orange), compared against ground truth (black dashed line).
    (c) Bland–Altman analysis across all subjects ($N = 9$) comparing PITN-Full and AugPITN (B) predictions.
    (d) Pearson correlation analysis showing agreement between predicted and true values for both models.
    }
    \label{fig:main_perf_comp}
\end{figure*}

\section{Experiment Design}
\subsection{Data}
\textcolor{black}{We used the \textit{demo }dataset from the PITN-Full paper,
which includes high-fidelity, cuffless blood pressure record-
ings acquired from wearable bioimpedance (BioZ) sensors.
While this dataset provides proof-of-concept validation, it is
limited in scope with only N = 9 subjects, which constrains
the generalizability of our findings and requires careful in-
terpretation of results. The physiological signals were first
scaled to normalize their amplitudes. Discontinuous sam-
ples were ignored to ensure continuity in the input time-
series data, which is critical for stable model training and
performance. \newline
For train/test splitting, the PITN’s approach of selecting one data point persampling approach was used, where a single data point was randomly selected from each pressure bin. This ensured broad bin coverage, but resulted in a sparse training set.}

\subsection{Baselines}
We compare our models against the following baselines for cuffless blood pressure estimation:

\begin{itemize}
  \item \textbf{PITN-Full~\cite{wang2024adversarial}}: Full Physics-Informed Temporal Network with temporal modeling, Taylor-based physics constraints, adversarial training (PGD), and contrastive learning—serving as the state-of-the-art benchmark.
  
  \item \textbf{PITN-Base~\cite{wang2024adversarial}}:Reduced version of PITN-Full with temporal modeling and physics-informed loss; excludes adversarial and contrastive components to isolate core effects.
  
  \item \textbf{PITN-Base (no Physics)}: PITN-Base without the physics loss term, used to assess the impact of physics-informed regularization.
  
  \item \textbf{DeepONet~\cite{lu2019deeponet}}: Standard Deep Operator Network using branch and trunk networks; no physics-based constraints.
  
  \item \textbf{PI-DeepONet}: DeepONet with Taylor-based physics constraints, allowing direct comparison between operator learning and temporal modeling.
\end{itemize}

\vspace{0.2cm}

\section{Results and Discussion}
\subsection{Performance Comparison}
In Fig.~\ref{fig:main_perf_comp}, the near-complete overlap of data points in the Bland--Altman analysis (c) and Pearson correlation plot (d) reinforces the strong agreement between AugPITN~(B) and PITN-Full predictions, visually confirming that the proposed operator-model based approach can closely replicate the behavior of a much more complex architecture. This is further supported by the numerical metrics in Table~\ref{tab:results}, where AugPITN~(B) achieves an identical correlation of $0.77$ and a comparable RMSE of $4.5$, versus PITN-Full’s RMSE of $4.45$. AugPITN~(A) also performs competitively, with a correlation of $0.77$ and RMSE of $4.58$, demonstrating that operator-guided supervision can replace adversarial and contrastive learning components while maintaining high predictive accuracy.

Table~\ref{tab:results} also reveals broader trends across benchmark models. PITN-Base provides a strong baseline (Corr: $0.73$, RMSE: $4.70$), while removing physics components leads to a substantial performance drop (Corr: $0.41$, RMSE: $8.26$), underscoring the importance of domain knowledge. Among operator models, PI-DeepONet outperforms DeepONet (Corr: $0.68$ vs. $0.58$), validating the utility of physiological constraints. Both AugPITN variants outperform all standalone models, confirming that pretrained neural operator supervision effectively transfers domain knowledge, enabling simplified models to rival or exceed more complex architectures.

\begin{table*}[ht]
    \centering
    \caption{Performance comparison across different models for BP estimation.}
    \label{tab:results}
    \resizebox{\textwidth}{!}{%
    \begin{tabular}{|c|c|c|c|c|c|c|c|}
        \hline
        \multirow{2}{*}{\textbf{Metrics}} & \multicolumn{7}{c|}{\textbf{Experimented Models}} \\
        \cline{2-8}
         & \textbf{PITN-Full} \cite{wang2024adversarial} 
         & \textbf{PITN-Base} 
         & \makecell{\textbf{PITN-Base} \\ \textbf{(no Physics)}} 
         & \textbf{DeepONet} \cite{lu2019deeponet} 
         & \textbf{PI-DeepONet} 
         & \makecell{\textbf{AugPITN (A)}  \\ \textbf{(PI-DeepONet, $\beta = 1.0$)}}
         & \makecell{\textbf{AugPITN (B)}  \\ \textbf{(PI-DeepONet, $\beta = 1.0$)}} \\
        \hline
        \textbf{Corr}  & 0.766& 0.725& 0.413& 0.579& 0.678& 0.768& \textbf{0.770}\\
        \textbf{RMSE}  & 4.452& 4.696& 8.263& 6.603& 5.510& 4.575& \textbf{4.501}\\
        \hline
    \end{tabular}%
    }
\end{table*}

\begin{table}[ht]
    \centering
    \caption{Ablation study of OSM regularization coefficient $\beta$ for AugPITN (A) and AugPITN (B) models. Metrics reported: correlation (Corr) and root-mean-squared error (RMSE).}
    \label{tab:ablation_beta}
    \resizebox{0.6\columnwidth}{!}{%
    \begin{tabular}{|c|c|c|c|}
        \hline
        \textbf{Model} & \boldmath{$\beta$} & \textbf{Corr} & \textbf{RMSE} \\
        \hline
        \multirow{4}{*}{AugPITN (A)} 
            & 0.05 &  0.709&  4.874\\
            & 0.10 &  0.710&  4.874\\
            & 0.50 &  0.710&  4.865\\
            & \textbf{1.00} & \textbf{0.768} &  \textbf{4.575}\\
        \hline
        \multirow{4}{*}{AugPITN (B)} 
            & 0.05 &  0.638&  5.641\\
            & 0.10 &  0.657&  5.441\\
            & 0.50 &  0.704&  5.069\\
            & \textbf{1.00} &  \textbf{0.770} & \textbf{4.501} \\
        \hline
    \end{tabular}%
    }
\end{table}

\subsection{Physics-Informed Regularization and Ablation }
\subsubsection{ DeepONet vs. PI-DeepONet}
The physics-informed loss significantly enhances operator learning in hemodynamic modeling. Standard DeepONet, without regularization, achieved a correlation of 0.40 and RMSE of 8.02, underperforming compared to the baseline \textbf{PITN} model. Incorporating Taylor-based constraints in \textbf{PI-DeepONet} improved performance to a correlation of 0.62 and RMSE of 6.24. The Taylor expansion enforces temporal consistency and physiological plausibility, guiding the model to realistic blood pressure transitions. Like in \textbf{PITN-Base}, it compensates for limited data by embedding physical priors. The regularization term $\lambda_{\text{physics}}$ balances learning accuracy with physiological fidelity, making \textbf{PI-DeepONet} more effective for personalized cardiovascular modeling where training data is scarce.

\subsubsection{OSM Regularization Coefficient $\beta$ Ablation}
Table~\ref{tab:ablation_beta} summarizes the impact of varying $\beta$ in AugPITN, which augments PITN-Base using a frozen neural operator (e.g., DeepONet or PI-DeepONet) as a supervisor. AugPITN~(A) compares the operator's outputs to ground truth (\textit{target-referenced supervision}), while AugPITN~(B) aligns PITN predictions with operator outputs (\textit{prediction alignment}). 
The results reveal that increasing $\beta$, which governs the influence of operator-based supervision, generally improves performance in both approaches. At $\beta = 1.0$, both AugPITN~(A) and AugPITN~(B) achieve peak performance: AugPITN~(A) reaches $\mathrm{Corr} \approx 0.768$ and $\mathrm{RMSE} \approx 4.575$, while AugPITN~(B) attains $\mathrm{Corr} \approx 0.770$ and $\mathrm{RMSE} \approx 4.501$.

These findings demonstrate that stronger supervisory signals from the operator model lead to improved generalization and predictive accuracy. Notably, AugPITN~(B) shows greater sensitivity to $\beta$, with performance substantially deteriorating at lower values ($\beta = 0.05$--$0.10$), suggesting that effective alignment with the operator requires sufficient regularization strength. In contrast, AugPITN~(A) exhibits more stable performance across a wider range of $\beta$ values, potentially due to its direct grounding in true labels. Overall, these results highlight the efficacy of neural operator-based supervision in enhancing time-series forecasting models, with distinct advantages depending on whether supervision is grounded in data or in learned operator behavior.

\subsection{AugPITN: Efficiency and Simplification}
\textbf{Complexity Overheads in PITN-Full :}
The implementation of adversarial and contrastive learning components in \textbf{PITN-Full} introduces substantial complexity in both preprocessing and architectural design, as illustrated in Figure~\ref{fig:PITN-Full}. Adversarial training requires generating perturbed samples through Projected Gradient Descent (PGD), necessitating careful management of dual data flows for both clean and adversarial inputs. The contrastive learning module demands sophisticated positive and negative pair generation based on physiological similarity thresholds, along with complex batch composition strategies to ensure adequate representation during training. These components require auxiliary layer normalizations that must be conditionally disabled during inference to account for distributional differences between clean and adversarial samples.

\textbf{Hyperparameter Sensitivity and Tuning Burden :}
The hyperparameter landscape of \textbf{PITN-Full} becomes particularly challenging, requiring optimization of at least eight critical parameters, including adversarial perturbation strength ($\eta = 0.2$), contrastive temperature ($\tau$), positive pair threshold ($y_{\text{shift}} = 2$), physics loss weighting ($\gamma = 1$), and perturbation bounds ($\epsilon = 0.2$). The paper's parameter sensitivity analysis (in Figure 8, see \cite{wang2024adversarial} ) demonstrates that performance fluctuates significantly with varying $y_{\text{shift}}$ values, while correlation degrades substantially as $\gamma$ increases beyond 1. Each parameter exhibits complex interdependencies, making convergence sensitive to precise tuning and often requiring extensive grid searches, as evidenced by the multiple $\beta$ configurations (explored in Table II, see \cite{wang2024adversarial}) for regularization strength.

\textbf{Simplification via Operator Supervision:}
In contrast, the \textbf{OSM supervision} approach dramatically simplifies this complexity, as shown in Figure~\ref{fig:arch}(c). The operator model can be pre-trained once on high-fidelity physiological data using standard supervised learning techniques, requiring only the standard physics-informed loss weighting. Once frozen, this pre-trained model serves as a reliable supervisor, reducing the entire augmentation framework to a single additional hyperparameter ($\beta$) controlling supervision strength. This architectural simplification eliminates the need for adversarial sample generation, contrastive pair management, and multi-component loss balancing while maintaining equivalent predictive performance, as demonstrated by the identical correlation scores (0.77) achieved by both PITN-Full and AugPITN variants in Table~\ref{tab:results}.

\textbf{Performance Parity with Reduced Cost :}
AugPITN (B) matches the predictive performance of the more complex PITN-Full model while substantially simplifying the training pipeline. It retains a comparable parameter count, yet reduces training time by approximately 4\%, highlighting improved efficiency. This performance parity is attained without relying on PITN-Full’s adversarial and contrastive learning components, indicating that AugPITN (B) can match predictive accuracy through a simpler training setup.

\section{Conclusion and Future Work}
This work demonstrates that physics-informed neural operators can effectively supervise simplified architectures for hemodynamic modeling. We show that operator-based knowledge distillation successfully replaces complex adversarial and contrastive learning strategies while achieving equivalent performance (correlation: 0.77, RMSE: 4.5). By using pretrained PI-DeepONet model as frozen supervisors, we reduce the training framework to a single primary hyperparameter while maintaining performance parity and dramatically simplifying implementation. The operator supervision paradigm offers a scalable approach for physics-informed machine learning in healthcare, enabling practitioners to leverage complex model benefits without implementation burden.\newline
\textcolor{black}{Although our evaluation highlights promising efficiency gains, the absence of public benchmark validation underscores the need for future research, paving the way for broader investigation and development. Future work should also explore operator transferability across physiological systems, integration of multi-modal sensing, and theoretical foundations for optimal supervision strategies, building on our core finding that operator supervision can replace complex adversarial training in physics-informed physiological modeling.}

\bibliographystyle{IEEEtran}
\bibliography{reference}

\begin{thebibliography}{10}
\providecommand{\url}[1]{#1}
\csname url@samestyle\endcsname
\providecommand{\newblock}{\relax}
\providecommand{\bibinfo}[2]{#2}
\providecommand{\BIBentrySTDinterwordspacing}{\spaceskip=0pt\relax}
\providecommand{\BIBentryALTinterwordstretchfactor}{4}
\providecommand{\BIBentryALTinterwordspacing}{\spaceskip=\fontdimen2\font plus
\BIBentryALTinterwordstretchfactor\fontdimen3\font minus \fontdimen4\font\relax}
\providecommand{\BIBforeignlanguage}[2]{{%
\expandafter\ifx\csname l@#1\endcsname\relax
\typeout{** WARNING: IEEEtran.bst: No hyphenation pattern has been}%
\typeout{** loaded for the language `#1'. Using the pattern for}%
\typeout{** the default language instead.}%
\else
\language=\csname l@#1\endcsname
\fi
#2}}
\providecommand{\BIBdecl}{\relax}
\BIBdecl

\bibitem{Wang2023}
S.~Wang and P.~Perdikaris, ``Long-time integration of parametric evolution equations with physics-informed deeponets,'' \emph{Journal of Computational Physics}, vol. 475, 2 2023.

\bibitem{Brecht2024}
R.~Brecht and A.~Bihlo, ``M-eniac: A physics-informed machine learning recreation of the first successful numerical weather forecasts,'' \emph{Geophysical Research Letters}, vol.~51, 5 2024.

\bibitem{Lu2021}
L.~Lu, P.~Jin, G.~Pang, Z.~Zhang, and G.~E. Karniadakis, ``Learning nonlinear operators via deeponet based on the universal approximation theorem of operators,'' \emph{Nature Machine Intelligence}, vol.~3, pp. 218--229, 3 2021.

\bibitem{Wang2021}
\BIBentryALTinterwordspacing
S.~Wang, H.~Wang, and P.~Perdikaris, ``Learning the solution operator of parametric partial differential equations with physics-informed deeponets,'' Tech. Rep., 2021. [Online]. Available: \url{https://www.science.org}
\BIBentrySTDinterwordspacing

\bibitem{wang2024adversarial}
R.~Wang, M.~Qi, Y.~Shao, A.~Zhou, and H.~Ma, ``Adversarial contrastive learning based physics-informed temporal networks for cuffless blood pressure estimation,'' \emph{arXiv preprint arXiv:2408.08488}, 2024.

\bibitem{banerjee2024physics}
C.~Banerjee, K.~Nguyen, C.~Fookes, and K.~George, ``Physics-informed computer vision: A review and perspectives,'' \emph{ACM Computing Surveys}, vol.~57, no.~1, pp. 1--38, 2024.

\bibitem{banerjee2025survey}
C.~Banerjee, K.~Nguyen, C.~Fookes, and M.~Raissi, ``A survey on physics informed reinforcement learning: Review and open problems,'' \emph{Expert Systems with Applications}, p. 128166, 2025.

\bibitem{banerjee2024pinns}
C.~Banerjee, K.~Nguyen, O.~Salvado, T.~Tran, and C.~Fookes, ``Pinns for medical image analysis: A survey,'' \emph{arXiv preprint arXiv:2408.01026}, 2024.

\bibitem{kissas2020machine}
G.~Kissas, Y.~Yang, E.~Hwuang, W.~R. Witschey, J.~A. Detre, and P.~Perdikaris, ``Machine learning in cardiovascular flows modeling: Predicting arterial blood pressure from non-invasive 4d flow mri data using physics-informed neural networks,'' \emph{Computer Methods in Applied Mechanics and Engineering}, vol. 358, p. 112623, 2020.

\bibitem{herrero2022ep}
C.~Herrero~Martin, A.~Oved, R.~A. Chowdhury, E.~Ullmann, N.~S. Peters, A.~A. Bharath, and M.~Varela, ``Ep-pinns: Cardiac electrophysiology characterisation using physics-informed neural networks,'' \emph{Frontiers in Cardiovascular Medicine}, vol.~8, p. 768419, 2022.

\bibitem{Cardoso-Bihlo2023}
\BIBentryALTinterwordspacing
E.~Cardoso-Bihlo and A.~Bihlo, ``Exactly conservative physics-informed neural networks and deep operator networks for dynamical systems,'' 11 2023. [Online]. Available: \url{http://arxiv.org/abs/2311.14131}
\BIBentrySTDinterwordspacing

\bibitem{DeRyck2022}
T.~D. Ryck and S.~Mishra, ``Error analysis for physics-informed neural networks (pinns) approximating kolmogorov pdes,'' \emph{Advances in Computational Mathematics}, vol.~48, 12 2022.

\bibitem{Ernst2024}
L.~Ernst and K.~Urban, ``A certified wavelet-based physics-informed neural network for the solution of parameterized partial differential equations,'' \emph{IMA Journal of Numerical Analysis}, 2 2024.

\bibitem{Sharma2023}
P.~Sharma, L.~Evans, M.~Tindall, and P.~Nithiarasu, ``Stiff-pdes and physics-informed neural networks,'' pp. 2929--2958, 6 2023.

\bibitem{Taebi2022}
A.~Taebi, ``Deep learning for computational hemodynamics: A brief review of recent advances,'' \emph{Fluids}, vol.~7, 6 2022.

\bibitem{Xie2025}
H.~Xie, X.~Zhao, N.~Zhang, J.~Liu, G.~Yang, Y.~Cao, J.~Xu, L.~Xu, Z.~Sun, Z.~Wen, S.~Chai, and D.~Liu, ``Machine learning–based hemodynamics quantitative assessment of pulmonary circulation using computed tomographic pulmonary angiography,'' \emph{International Journal of Cardiology}, vol. 437, 10 2025.

\bibitem{Zhao2023}
L.~Zhao, C.~Liang, Y.~Huang, G.~Zhou, Y.~Xiao, N.~Ji, Y.~T. Zhang, and N.~Zhao, ``Emerging sensing and modeling technologies for wearable and cuffless blood pressure monitoring,'' \emph{npj Digital Medicine}, vol.~6, 12 2023.

\bibitem{Hua2024}
J.~Hua, M.~Su, J.~Wu, Y.~Zhou, Y.~Guo, Y.~Shi, and L.~Pan, ``Wearable cuffless blood pressure monitoring: From flexible electronics to machine learning,'' \emph{Wearable Electronics}, vol.~1, pp. 78--90, 12 2024.

\bibitem{Shen2025}
J.~Shen, J.~Wu, H.~Liang, Z.~Zhao, K.~Li, K.~Zhu, K.~Wang, Y.~Ma, W.~Hu, C.~Guo, Y.~Zhang, and B.~Hu, ``Physiological signal analysis using explainable artificial intelligence: A systematic review,'' \emph{Neurocomputing}, vol. 618, 2 2025.

\bibitem{Sel2024}
K.~Sel, D.~Osman, F.~Zare, S.~M. Shahrbabak, L.~Brattain, J.~Hahn, O.~T. Inan, R.~Mukkamala, J.~Palmer, D.~Paydarfar, R.~I. Pettigrew, A.~A. Quyyumi, B.~Telfer, and R.~Jafari, ``Building digital twins for cardiovascular health: From principles to clinical impact,'' \emph{Journal of the American Heart Association}, 10 2024.

\bibitem{kovachki2023neural}
N.~Kovachki, Z.~Li, B.~Liu, K.~Azizzadenesheli, K.~Bhattacharya, A.~Stuart, and A.~Anandkumar, ``Neural operator: Learning maps between function spaces with applications to pdes,'' \emph{Journal of Machine Learning Research}, vol.~24, no.~89, pp. 1--97, 2023.

\bibitem{li2024physics}
Z.~Li, H.~Zheng, N.~Kovachki, D.~Jin, H.~Chen, B.~Liu, K.~Azizzadenesheli, and A.~Anandkumar, ``Physics-informed neural operator for learning partial differential equations,'' \emph{ACM/JMS Journal of Data Science}, vol.~1, no.~3, pp. 1--27, 2024.

\bibitem{lu2019deeponet}
L.~Lu, P.~Jin, and G.~E. Karniadakis, ``Deeponet: Learning nonlinear operators for identifying differential equations based on the universal approximation theorem of operators,'' \emph{arXiv preprint arXiv:1910.03193}, 2019.

\end{thebibliography}

\end{document}